\documentclass[runningheads]{llncs}

\usepackage{eccv}

\usepackage{eccvabbrv}

\usepackage{graphicx}
\usepackage{booktabs}
\usepackage{multirow}
\usepackage{makecell}
\usepackage{pifont}

\usepackage[accsupp]{axessibility}  %

\usepackage{hyperref}

\usepackage{orcidlink}

\newcommand{\cmark}{\ding{51}}

\begin{document}

\title{Text-Guided 6D Object Pose Rearrangement via Closed-Loop VLM Agents}

\titlerunning{Text-Guided 6D Object Pose Rearrangement via Closed-Loop VLM Agents}
\author{Sangwon Baik\inst{1} \and
Gunhee Kim\inst{1} \and
Mingi Choi\inst{1} \and
Hanbyul Joo\inst{1, 2}}

\authorrunning{S.~Baik et al.}

\institute{Seoul National University \and
RLWRLD \\
\url{https://tlb-miss.github.io/vlmpose}}

\maketitle

\begin{figure}
    \includegraphics[width=\linewidth, trim={0cm 0cm 0cm 0cm},clip]{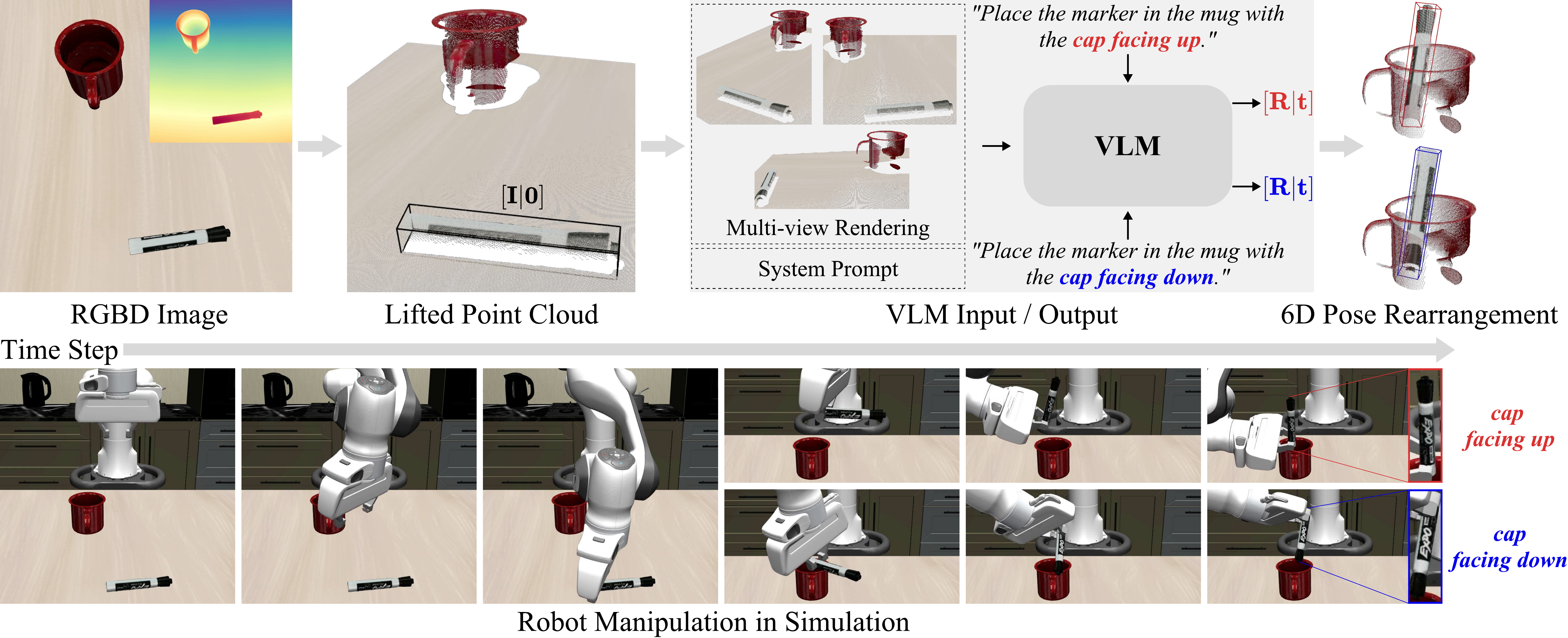}
    \caption{Given a 3D scene and a text instruction specifying the desired goal state of the scene, our method uses a VLM to iteratively refine the target object's 6D pose in a closed loop until it reaches the goal.}
    \label{fig:teaser}
\end{figure}
\begin{abstract}
Vision-Language Models (VLMs) exhibit strong visual reasoning capabilities, yet they still struggle with 3D understanding. In particular, VLMs often fail to infer a text-consistent goal 6D pose of a target object in a 3D scene.
However, we find that with some inference-time techniques and iterative reasoning, VLMs can achieve dramatic performance gains. Concretely, given a 3D scene represented by an RGB-D image (or a compositional scene of 3D meshes) and a text instruction specifying a desired state change, we repeat the following loop: observe the current scene; evaluate whether it is faithful to the instruction; propose a pose update for the target object; apply the update; and render the updated scene. Through this closed-loop interaction, the VLM effectively acts as an agent.
We further introduce three inference-time techniques that are essential to this closed-loop process: (i) multi-view reasoning with supporting view selection, (ii) object-centered coordinate system visualization, and (iii) single-axis rotation prediction. Without any additional fine-tuning or new modules, our approach surpasses prior methods at predicting the text-guided goal 6D pose of the target object. It works consistently across both closed-source and open-source VLMs. Moreover, when combining our 6D pose prediction with simple robot motion planning, it enables more successful robot manipulation than recent Vision-Language-Action models (VLAs). Finally, we conduct an ablation study to demonstrate the necessity of each proposed technique.
\end{abstract}

\section{Introduction}
\label{sec:intro}

Modern Vision-Language Models (VLMs) are brilliant observers that can accurately describe the detailed patterns of a key, yet they remain limited in spatial reasoning when tasked with fitting that key into a lock. While these models show human-level fluency in interpreting complex scenes, they still struggle to translate a language instruction such as \textit{``Place the marker in the mug with the cap facing up.''} into a precise goal 6D pose. This spatial gap becomes more severe when VLMs infer in a feed-forward manner from a single image. These one-shot systems lack the visual feedback necessary to resolve depth ambiguities, occlusions, and complex geometric constraints.

In this work, we address this limitation by moving from passive one-shot prediction to active closed-loop refinement. We introduce a training-free framework for text-guided 6D pose rearrangement, which we formulate as a sequential decision problem over rendered observations of a 3D scene.  At each iteration, the VLM alternates between an evaluator that judges whether the current scene is faithful to the instruction and a proposer that predicts an incremental 6D pose update for the target object. In the closed-loop view of our framework, this pose update serves as the agent's action. This allows the model to leverage direct visual feedback from a 3D renderer to iteratively improve alignment and correct spatial errors over multiple steps. Unlike prior methods that rely on text-based abstractions, our approach keeps the VLM in a tighter visual loop and refines the object pose directly from rendered feedback.

To make this closed-loop process stable and precise, we introduce three inference-time techniques that provide stronger 3D grounding for the VLM. First, we use \textbf{multi-view reasoning with supporting view selection} to reduce the depth ambiguity of single-view observations and handle occlusions. Second, we propose \textbf{object-centered coordinate system visualization}, which renders explicit 3D axes on the target object and provides the VLM with clear cues about direction and relative scale. Third, we use \textbf{single-axis rotation prediction} to simplify the SO(3) search space by breaking rotational reasoning into axis-aligned steps. We further maintain a context memory of prior evaluations and predicted pose updates to reduce oscillating predictions and support stable convergence toward the goal state.

The contributions of this paper are as follows:
\begin{itemize}
    \item We introduce a closed-loop agentic framework for text-guided 6D pose rearrangement that operates entirely training-free, using pre-trained VLMs in alternating evaluator and proposer roles.
    \item We propose a set of inference-time 3D grounding techniques, including multi-view reasoning with supporting view selection, object-centered coordinate system visualization, and single-axis rotation prediction, which significantly improve the 3D spatial reasoning of VLMs. We further demonstrate the effectiveness of these techniques through ablation studies.
    \item Through extensive evaluation on the Open6DOR V2~\cite{sofar} and SIMPLER~\cite{simpler} benchmarks, we show that our method achieves substantial performance gains, particularly in orientation-sensitive tasks, and leads to higher success rates in zero-shot robotic manipulation.
\end{itemize}

\section{Related Work}
\label{sec:rel_work}

\subsection{Vision-Language Models as Agents}
\label{sec:vlm_agent}

Recent work has increasingly used VLMs as interactive agents that observe visual inputs, take actions, and update their behavior based on feedback. 
In GUI settings, CogAgent~\cite{cogagent} studies screenshot-based GUI understanding and navigation. 
ShowUI~\cite{showui} formulates GUI interaction as a vision-language-action problem for visual GUI agents. 
UI-TARS~\cite{ui-tars} and UI-TARS-2~\cite{ui-tars-2} further treat screenshot-based GUI interaction as a native agent problem with multi-step decision-making and iterative interaction. 
In web environments, SeeAct~\cite{seeact} studies grounded action generation for generalist web agents, while WebVoyager~\cite{webvoyager} demonstrates end-to-end multimodal web interaction on real websites. 
Beyond software interfaces, AVA~\cite{ava} applies a VLM-based agent to a game environment, and VIGA~\cite{viga} uses an iterative loop of generation, rendering, and verification in graphics and 3D settings.
Under this broader view of VLMs as agents, we study text-guided object 6D pose rearrangement, where a VLM iteratively refines the target pose through closed-loop interaction with the rendered scene.

\subsection{3D Spatial Reasoning in Vision-Language Models}
\label{sec:vlm_3d_reasoning}

Recent work improves VLM 3D reasoning via spatial supervision, geometric priors, and reconstruction. SpatialVLM~\cite{spatialvlm} and SpatialRGPT~\cite{spatialrgpt} utilize large-scale supervision and depth-aware modeling. Meanwhile, SpatialPIN~\cite{spatialpin} enables zero-shot reasoning via 3D foundation model priors, while Spatial-MLLM~\cite{spatial-mllm} leverages geometry-aware features from 2D observations. Others, like VLM-3R~\cite{vlm-3r} and G$^2$VLM~\cite{g2vlm}, unify spatial reasoning with 3D reconstruction. For manipulation, VoxPoser~\cite{voxposer} grounds language into 3D value maps, and RoboBrain~\cite{robobrain} integrates high-level planning with affordance perception.

In robotics, generalist policies like RT-1~\cite{rt-1} and Octo~\cite{octo} have evolved into vision-language-action (VLA) models such as RT-2~\cite{rt-2} and OpenVLA~\cite{openvla}. Recent advances further incorporate 3D representations~\cite{spatialvla, 3d-vla} and emerging foundation models~\cite{pi_0, gr00t}.

Specific to language-conditioned rearrangement, Dream2Real~\cite{dream2real} employs VLM-based scoring over rendered candidates, Open6DOR-GPT~\cite{open6dor-gpt} utilizes simulation assistance, and SoFar~\cite{sofar} relies on structured 6-DoF scene graphs. Unlike these methods that depend on candidate scoring or structured/simulator-assisted search, our approach keeps the VLM in a tighter visual loop, iteratively refining the target 6D pose directly from rendered visual feedback via inference-time techniques.

\section{Method}
\label{sec:method}

\begin{figure}[t]
    \includegraphics[width=\linewidth, trim={0cm 0cm 0cm 0cm},clip]{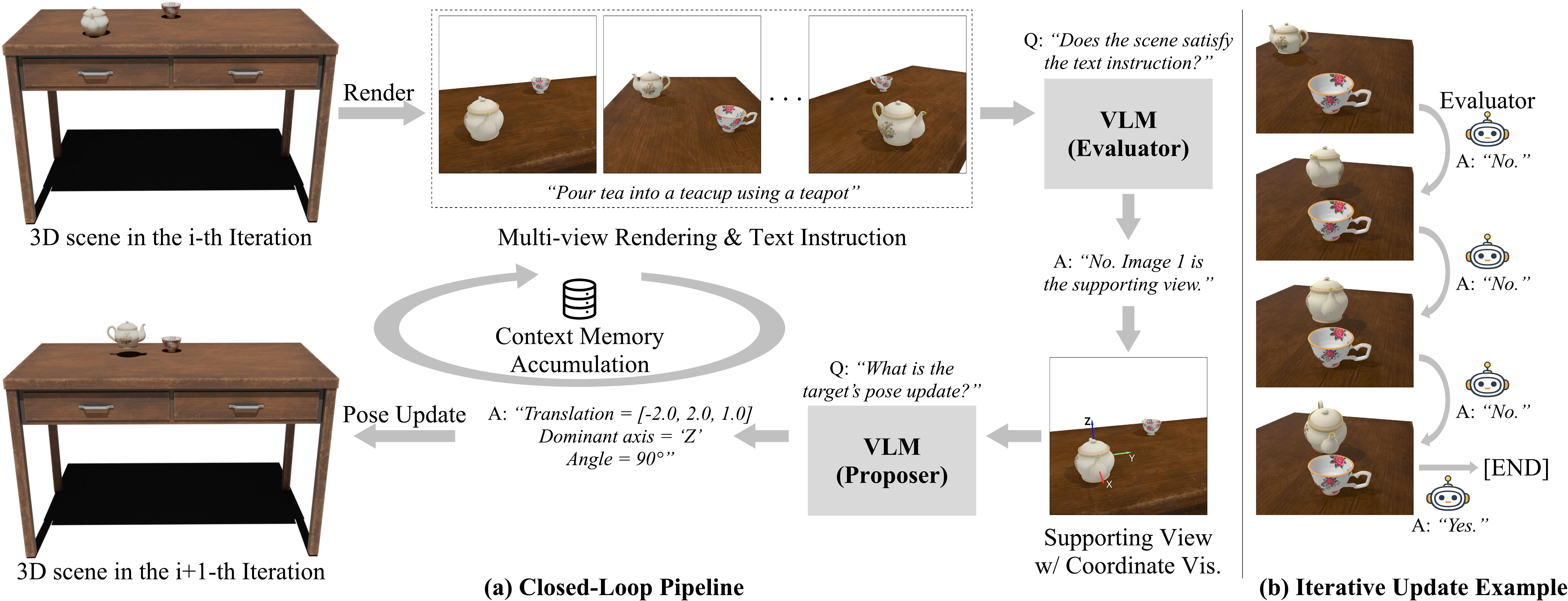}
    \caption{\textbf{Our Method Overview.} (a) We first ask the evaluator to assess how well the rendered multi-view images satisfy the input text instruction. If the current scene is judged inconsistent with the instruction, we provide the proposer with the supporting view, augmented with coordinate system visualization, that best explains the evaluator’s judgment. Based on this input, the proposer predicts an incremental 6D pose update. The predicted update is then applied to the target object, and this process is repeated iteratively. Throughout all iterations, both roles receive the accumulated context memory. (b) Example of iterative pose updates produced by our method for a scene containing a teacup and a teapot on a table, given the text instruction \textit{``Pour tea into a teacup using a teapot.''}}
    \label{fig:method_overview}
\end{figure}

Given a 3D scene represented either by an RGB-D image with camera intrinsics or by a composition of meshes, our goal is to rearrange the pose of a target object so that the scene satisfies a given text instruction $\mathbf{c}$. To this end, we propose a method that can be applied to pre-trained Vision-Language Models (VLMs) without introducing specialized modules or requiring additional fine-tuning. In our framework, a single VLM is used iteratively in two roles: an evaluator that measures how well the current scene aligns with the text instruction, and a proposer that predicts an incremental 6D pose update for the target object. This closed-loop process is supported by three inference-time techniques that improve the VLM's 3D reasoning performance. Fig.~\ref{fig:method_overview} summarizes the overall pipeline. For visual clarity, we show the mesh setting, although the same overall procedure is applied to the RGB-D setting after preprocessing.

\subsection{Closed-Loop Formulation}
\label{sec:closed_loop_formulation}

Let $\vec{\mathcal{S}}_i$ denote the 3D scene state at $i$-th iteration, represented either by object meshes or by point clouds lifted from an RGB-D observation.
Given an iteration-dependent camera set $\Pi_i = \{\pi_i^{(k)}\}_{k=1}^{K}$, the renderer $\mathcal{R}$ produces the multi-view images
\begin{equation}
    \vec{\mathcal{I}}_i = \mathcal{R}(\vec{\mathcal{S}}_i, \Pi_i)
    = \{\vec{I}_i^{(k)}\}_{k=1}^{K},
    \label{eq:render_obs}
\end{equation}
where $\vec{I}_i^{(k)} = \mathcal{R}(\vec{\mathcal{S}}_i, \pi_i^{(k)})$ is the rendered image of the $k$-th camera at iteration $i$. For each iteration, the evaluator $\Phi_{\mathrm{eval}}$ predicts
\begin{equation}
    (y_i, k_i, r_i^{\mathrm{eval}}) = \Phi_{\mathrm{eval}}(\vec{\mathcal{I}}_i, \mathbf{c}),
    \label{eq:evaluator}
\end{equation}
where $y_i \in \{0, 1\}$ indicates whether the current scene is faithful to the instruction $\mathbf{c}$, $k_i \in \{1,...,K\}$ is the supporting-view index identified as the best view among rendered views, and $r_i^{\mathrm{eval}}$ is the evaluator's rationale for selecting $\vec{I}_i^{(k_i)}$ as the supporting view. If $y_i = 0$, meaning that the current scene does not satisfy the instruction, the proposer $\Phi_{\mathrm{prop}}$ predicts an incremental 6D pose update for the target object:
\begin{equation}
    (\hat{\mathbf{t}}_i, \omega_i, \theta_i, r_i^{\mathrm{prop}}) = \Phi_{\mathrm{prop}}(\tilde{\vec{I}}_i^{(k_i)}, \mathbf{c}),
    \label{eq:proposer}
\end{equation}
where $\tilde{\vec{I}}_i^{(k_i)}$ denotes $\vec{I}_i^{(k_i)}$ augmented with the object-centered coordinate system visualization, $\hat{\mathbf{t}}_i \in \mathbb{R}^3$ is the translation in normalized axis units, $\omega_i \in \{x,y,z\}$ is the dominant rotation axis, $\theta_i \in \mathbb{R}$ is the rotation angle, and $r_i^{\mathrm{prop}}$ is the proposer's rationale for the target pose update. In the closed-loop view of our framework, $(\hat{\mathbf{t}}_i, \omega_i, \theta_i)$ is the agent's action. Applying this predicted pose update to the target object yields the next scene state $\vec{\mathcal{S}}_{i+1}$. For simplicity, context memory is omitted from Eqs.~\ref{eq:evaluator} and \ref{eq:proposer}.

\subsection{Overall Procedure with Inference-Time Techniques}
\label{sec:overview}

To focus only on the main objects relevant to the instruction $\mathbf{c}$, we first perform target and related object selection on the initial scene $\vec{\mathcal{S}}_0$ before starting the loop.
To do this, we compute an axis-aligned bounding box (AABB) that encloses all objects in the $\vec{\mathcal{S}}_0$, place cameras on a circular trajectory around its center, and render multi-view images together with per-object masks. From these images, we ask the VLM to select the view in which the objects are most clearly visible and distinguishable. Using the masks in the selected view, we annotate each object with a bounding box and label, and ask the VLM to identify the target object and the related objects mentioned in the text instruction. Fig.~\ref{fig:system_1} includes an example of the image provided to the VLM for this target and related object selection step. Now, we write the scene state $\vec{\mathcal{S}}_i$ more explicitly as follow:
\begin{equation}
    \vec{\mathcal{S}}_i = (\mathbf{O}^{tgt}, \mathbf{O}^{rel}, \mathbf{O}^{oth}, \mathbf{R}_i^{tgt}, \mathbf{t}_i^{tgt}),
    \label{eq:scene_formulation}
\end{equation}
where $\mathbf{O}^{tgt} \in \mathbb{R}^{N_{tgt} \times 3}$ is vertices of the target object, $\mathbf{O}^{rel} \in \mathbb{R}^{N_{rel} \times 3}$ represents vertices of the related objects, $\mathbf{O}^{oth} \in \mathbb{R}^{N_{oth} \times 3}$ represents vertices of the other objects in the scene, $\mathbf{R}_i^{tgt} \in SO(3)$ is the rotation matrix of the $\mathbf{O}^{tgt}$ in the world coordinate system, and $\mathbf{t}_i^{tgt} \in \mathbb{R}^3$ is the translation vector of the $\mathbf{O}^{tgt}$ in the world coordinate system. Note that $\mathbf{O}^{tgt}$ is defined in canonical coordinates, while $\mathbf{O}^{rel}$ and $\mathbf{O}^{oth}$ are defined in world coordinates. This notation setting is chosen because the 6D pose of $\mathbf{O}^{tgt}$ changes in each iteration, but the 6D poses of the other objects do not change. $\mathbf{O}^{tgt}_i = \mathbf{R}_i^{tgt}\mathbf{O}^{tgt} + \mathbf{t}_i^{tgt} \in \mathbb{R}^{N_{tgt} \times 3}$ is the target object defined in world coordinates.

After target and related object selection, our VLM agent enters the closed-loop inference phase. We introduce the following three techniques to improve the VLM's performance in the loop: (1) multi-view reasoning with supporting view selection; (2) object-centered coordinate system visualization; and (3) single-axis rotation prediction. Specifically, technique (1) is applied to the evaluator, technique (2) is always applied to the proposer and optionally to the evaluator when directional disambiguation is needed, and technique (3) is applied to the proposer. Fig.~\ref{fig:ablation_mesh} shows the necessity of each technique.

There are two challenges when inferring 6D pose rearrangement from a single-view image. First, due to depth ambiguity, it is difficult to judge whether the object is placed correctly along the forward (or backward) direction. Second, if one of the interacting objects becomes fully occluded in that view during inference, the required spatial reasoning can no longer be performed from that view alone. To address these issues, we introduce multi-view reasoning. At iteration $i$, $\Pi_i$ consists of $K$ equally spaced cameras placed on a circular trajectory such that the center of the AABB enclosing $\mathbf{O}^{tgt}_i$ and $\mathbf{O}^{rel}$ is the center of the frame, and all of these objects remain within the image frame. Now, multi-view images are rendered as in Eq.~\ref{eq:render_obs}, and the evaluator $\Phi_{\mathrm{eval}}$ reasons as in Eq.~\ref{eq:evaluator}. Concretely, $\Phi_{\mathrm{eval}}$ checks whether the target and related objects are present, whether the instruction-specified spatial relations are satisfied, and whether any physically implausible interpenetration is present. If the evaluator judges the current scene faithful, the loop terminates.

Otherwise, the proposer $\Phi_{\mathrm{prop}}$ predicts the pose update through Eq.~\ref{eq:proposer}. The proposer takes as input the supporting view augmented with object-centered coordinate system visualization, denoted by $\tilde{I}_i^{(k_i)}$. Coordinate system visualization provides explicit visual cues for direction and scale. This is particularly important for reasoning about rotations. We apply rotations according to the right-hand rule. However, if the coordinate system is described only in text without explicit axis visualization, even humans find it difficult to distinguish clockwise from counterclockwise rotation about a given axis. It also disambiguates directional concepts such as left, right, forward, and backward across views. For instance, a direction that appears as left in the front view corresponds to right in the back view, which can easily lead the VLM to incorrect judgments. By visually associating each axis with a direction (e.g., +x is front, +y is right, and +z is top), the VLM can reason about directions consistently across views. For such direction-sensitive tasks, $\Phi_{\mathrm{eval}}$ can also be given these augmented multi-view images. In addition, by assuming that each axis has unit length, the VLM can infer translations independently of the absolute scale of the scene. The coordinate system is visualized in an object-centered manner. At the $i$-th iteration, the system's origin is set at the center of the $\mathbf{O}^{tgt}_i$'s AABB, and the frame shares its axis directions with the world coordinate system. Each axis is drawn outward from the center of the corresponding AABB face with a prescribed length. In most cases, it is sufficient to choose this axis length $L$ such that:
\begin{equation}
    \label{eq:axis_length}
    \begin{gathered}
    L = \begin{cases}
                    \min(B_x, B_y, B_z), & 2\min(B_x, B_y, B_z) \leq \max(B_x, B_y, B_z), \\
                    \min(B_x, B_y, B_z)/2, & \text{otherwise}
                 \end{cases},
    \end{gathered}
\end{equation}
where $B_x$, $B_y$, and $B_z$ are the edge lengths of the $\mathbf{O}^{tgt}$'s (not $\mathbf{O}^{tgt}_i$'s) AABB, respectively. Note that the center of the coordinate system changes with each iteration, but the length of the axis is fixed. The VLM is instructed to assume a unit axis length, regardless of the actual axis length. Thus, the translational component of the proposer's output is converted to world coordinates as $\mathbf{t}_i = L \hat{\mathbf{t}}_i$.

Inferring an object’s full 3D rotation from an image is a challenging problem even for specialized models~\cite{Foundationpose, Foundpose, Gigapose, Picopose, SAM-6D} for object 6D pose estimation. In our setting, the ambiguity is even greater, since the model needs to infer the pose of the goal state rather than the current state. Therefore, directly inferring the full 3D rotation is nearly impossible for current pre-trained VLMs. However, as shown in Fig.~\ref{fig:method_overview}, we observe that restricting rotation inference to a single axis allows VLMs to perform reasonably well in combination with iterative reasoning. 
Because the coordinate system is object-centered but shares its axis directions with the world coordinate system, the predicted pose update is interpreted in the same axis convention. Thus, the rotational component of the $\Phi_{\mathrm{prop}}$'s output is interpreted as $\mathbf{R}_i = R_{\omega_i}(\theta_i)$ in world coordinates. Now, integrating all of the above, the scene state at the $(i+1)$-th iteration is updated as follows:
\begin{equation}
    \label{eq:scene_state_update}
    \begin{gathered}
    \mathbf{R}_{i+1}^{tgt} = \mathbf{R}_i\mathbf{R}_{i}^{tgt}, \;\; \mathbf{R}_i = R_{\omega_i}(\theta_i), \\
    \mathbf{t}_{i+1}^{tgt} = \mathbf{R}_i(\mathbf{t}_{i}^{tgt} - b_i) + b_i + \mathbf{t}_i, \;\; \mathbf{t}_i = L \hat{\mathbf{t}}_i, \\
    \mathbf{O}^{tgt}_{i+1} = \mathbf{R}_i(\mathbf{O}^{tgt}_{i} - b_i) + b_i + \mathbf{t}_i,
    \end{gathered}
\end{equation}
where $b_i$ is the center of the $\mathbf{O}^{tgt}_{i}$'s AABB.

Throughout the loop, the context memory accumulates prior evaluator judgments, previously predicted pose updates, and rationales. This memory is provided to both roles in subsequent iterations. This design improves overall loop performance, and its effect is analyzed through an ablation study in Sec.~\ref{sec:ablation}. The loop terminates when either the evaluator declares the scene faithful or the maximum number of iterations is reached. For an analysis of the maximum number of iterations, see Sec.~\ref{sec:max_iter_abl}. 

In the RGB-D setting, the only change is the preprocessing stage: we first lift the RGB-D input to a point cloud using the camera intrinsics, obtain object masks with off-the-shelf segmentation models~\cite{florence-2, sam, sam2, sam3}, and remove outlier points from each segmented object point cloud.

\section{Experiments}
\label{sec:exp}

We evaluate our framework from three perspectives. First, we measure how accurately it predicts text-guided 6D object goal poses on Open6DOR V2~\cite{sofar, open6dor-gpt}. Second, we test whether the predicted poses are useful for downstream robotic manipulation on Open6DOR V2 and SIMPLER~\cite{simpler}. Finally, we analyze the contribution of each component through ablation studies. For additional experiments on the effects of iterative loops, failure cases, and other analyses, see the supplementary material.

\begin{figure}[t]
    \includegraphics[width=\linewidth, trim={0cm 0cm 0cm 0cm},clip]{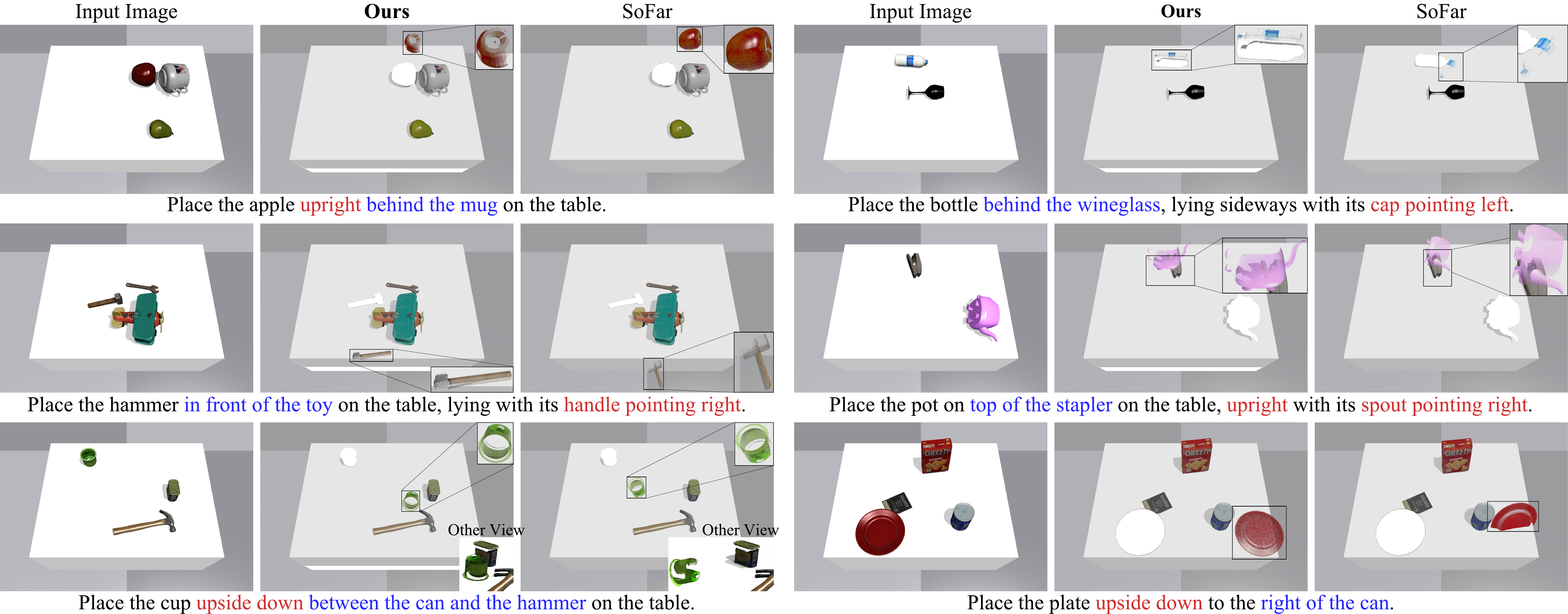}
    \caption{\textbf{Qualitative Comparison in the 6-DoF Rearrangement Task of the Open6DOR V2 Benchmark.} SoFar predicts positions reasonably well but often fails on rotation, while our method predicts both position and orientation more accurately. In the prompt, the target object's goal orientation is highlighted in red, and its goal position is highlighted in blue.}
    \label{fig:open6dor_6dof}
\end{figure}

\begin{table*}[t]
    \centering
    \small
    \setlength{\tabcolsep}{6pt}
    \renewcommand{\arraystretch}{1.15}
    \resizebox{\textwidth}{!}{%
    \begin{tabular}{llcccccccccc}
    \toprule
    \multicolumn{2}{c}{Method}
    & \multicolumn{3}{c}{Position Track} 
    & \multicolumn{4}{c}{Rotation Track} 
    & \multicolumn{3}{c}{6-DoF Track} \\
    \cmidrule(lr){1-2} \cmidrule(lr){3-5} \cmidrule(lr){6-9} \cmidrule(lr){10-12}
    & & Level 0 & Level 1 & Overall
    & Level 0 & Level 1 & Level 2 & Overall
    & Position & Rotation & Overall \\
    \midrule
    \multicolumn{2}{l}{Open6DOR-GPT~\cite{open6dor-gpt}}  & 56.7 & 78.6 & 61.0 & N/A & N/A & N/A & N/A & N/A & N/A & N/A \\
    \multicolumn{2}{l}{SoFar~\cite{sofar}}                & 96.5 & \textbf{82.0} & 93.7 & 37.2 & 22.3 & 42.4 & 31.3 & \textbf{94.7} & 28.0 & 26.3 \\
    \midrule
    \multirow{3}{*}{\centering \textbf{Ours}}
        & GPT-5.2~\cite{chatgpt}            & \textbf{97.5} & 81.6 & \textbf{94.4} & \textbf{63.3} & \textbf{50.1} & \textbf{58.3} & \textbf{56.6} & 91.0 & \textbf{51.1} & \textbf{46.4} \\
        & Gemini-2.5-Flash Lite~\cite{gemini2_5}   & 95.6 & 78.8 & 92.3 & 45.4 & 16.1 & 45.8 & 32.3 & 83.7 & 32.4 & 26.7 \\
        & InternVL3-14B~\cite{internvl3}    & 83.0 & 72.9 & 81.0 & 21.8 & 20.4 & 41.7 & 24.4 & 80.2 & 28.2 & 22.8 \\
    \bottomrule
    \end{tabular}%
    }
    \caption{\textbf{Quantitative Comparison in the 6-DoF Rearrangement Task of the Open6DOR V2 Benchmark.} We report success rates (\%) on the position, rotation, and 6-DoF tracks. Our method achieves comparable position accuracy to baselines while substantially improving rotation prediction and overall 6-DoF rearrangement performance. Open6DOR-GPT is reported only on the position track because its released code supports only that track.}
    \label{tab:open6dor_6dof}
\end{table*}

\subsection{Object 6-DoF Rearrangement Evaluation on Open6DOR V2}
\label{sec:open6dor_6dof}

\noindent \textbf{Dataset.}
Following SoFar~\cite{sofar}, we evaluate object 6-DoF rearrangement on the Open6DOR V2 benchmark~\cite{sofar, open6dor-gpt}. It is the first benchmark for the task of inferring the goal 6D pose of a target object from natural language instructions that specify object interactions within a scene. Open6DOR V2 consists of three tracks: the position track, the rotation track, and the 6D pose track. Each track covers a wide range of settings, from simple to complex, with higher levels corresponding to more challenging tasks. For example, in the position track, level 1 includes simple directional relations such as left, right, front, and back, whereas level 2 includes more challenging tasks such as between and center. In total, the benchmark contains 4,389 tasks. After excluding invalid scenes in which preprocessing failed, we conduct experiments on the remaining 4,370 tasks.

\noindent \textbf{Metrics.}
Following SoFar, we report success rate as the evaluation metric. For the position and rotation tracks, a prediction is considered successful if it falls within a predefined tolerance. For the 6-DoF track, a task is counted as successful only when both the position and rotation predictions are successful.

\noindent \textbf{Baselines.}
We compare our method with Open6DOR-GPT~\cite{open6dor-gpt} and SoFar~\cite{sofar}. Like our method, both Open6DOR-GPT and SoFar use VLMs to infer the goal 6D pose of the target object. However, they rely more heavily on structured textual scene information, such as descriptions of coordinate axes and object bounding boxes, rather than directly refining poses through iterative visual feedback.
Furthermore, to demonstrate the generality of our method, we apply it to multiple VLMs, including GPT-5.2~\cite{chatgpt}, Gemini-2.5-Flash Lite~\cite{gemini2_5}, and InternVL3~\cite{internvl3}. These models also span different reasoning capabilities, including a reasoning model (GPT-5.2) and weaker- or non-reasoning models (Gemini-2.5-Flash Lite and InternVL3). For fair comparison, we use GPT-5.2 as the underlying VLM for both SoFar and Open6DOR-GPT.

\noindent \textbf{Results.}
Table.~\ref{tab:open6dor_6dof} shows the quantitative results of our method and the baselines on the Open6DOR V2 benchmark. Our method outperforms the baselines on almost all tasks. While SoFar performs comparably to our method on position prediction, it fails on rotation prediction in most cases.
For Open6DOR-GPT, only the code for the position track was released, so we report results only for that track. Gemini-2.5-Flash Lite achieves performance comparable to that of SoFar, and InternVL3, despite being a non-reasoning VLM, outperforms Open6DOR-GPT. Fig.~\ref{fig:open6dor_6dof} clearly highlights the performance gap between SoFar and our method. Because it relies primarily on a text scene graph describing the target and related objects, it tends to underutilize global visual cues from the scene. As a result, it may place objects outside the table or cause the target object to collide with objects that are not included in the scene graph, leading to physically invalid configurations.

\begin{figure}[!t]
    \includegraphics[width=\linewidth, trim={0cm 0cm 0cm 0cm},clip]{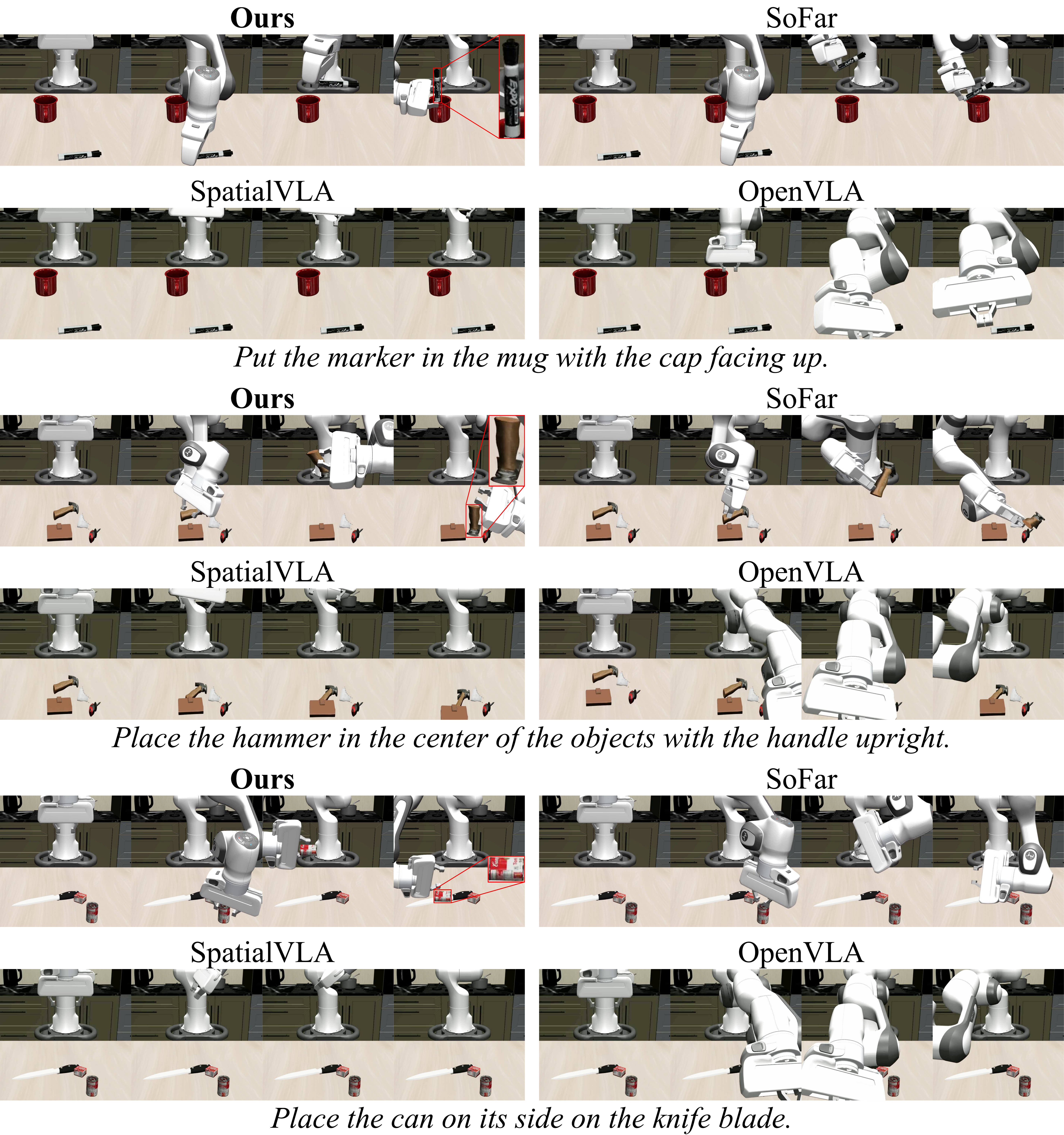}
    \caption{\textbf{Qualitative Comparison on Robot Manipulation Using Open6DOR V2 Objects and Scenes.} Our method predicts both position and orientation more accurately than the baselines.}
    \label{fig:libero}
\end{figure}

\subsection{Robot Manipulation Evaluation on Open6DOR V2}
\label{sec:open6dor_manip}

\noindent \textbf{Dataset and Metrics.}
We use the same task set and success metric as in Sec.~\ref{sec:open6dor_6dof}, and evaluate them under the LIBERO~\cite{libero} robot execution protocol with a Franka arm. Since Open6DOR V2 contains tasks in which the initial scene already matches the ground truth, we add an extra rule to prevent false positives: if the robot fails to grasp the target object and the scene remains unchanged, the trial is counted as a failure. We report results on 528 randomly sampled tasks.

\noindent \textbf{Baselines.}
We compare our method with SoFar and VLA methods, such as OpenVLA~\cite{openvla} and SpatialVLA~\cite{spatialvla}. We use Open X-Embodiment~\cite{oxe} pre-trained checkpoints for OpenVLA and SpatialVLA, rather than task-specific fine-tuned variants. Following SoFar, we adopt GraspNet~\cite{graspnet2, graspnet} for robot grasping and OMPL~\cite{ompl} for robot motion planning.
We additionally introduce two simple modifications in the LIBERO setting to improve grasping and motion planning. In the default setup, the robot wrist is aligned with the grasp point predicted by GraspNet. We instead align the gripper center with the predicted grasp point to obtain more robust grasps. We also modify the motion trajectory from the default trapezoidal lift--translate--lower path to a rectangular one: the robot first lifts the object, then moves in the $xy$-plane while adjusting the pose, and finally lowers it vertically at the goal. This reduces inter-object collisions and improves the success rate. We apply these modifications to both SoFar and our method for fair comparison.

\begin{table*}[t]
    \centering
    \small
    \setlength{\tabcolsep}{6pt}
    \renewcommand{\arraystretch}{1.15}
    \resizebox{\textwidth}{!}{%
    \begin{tabular}{lcccccccccc}
    \toprule
    \multirow{2}{*}{Method} 
    & \multicolumn{3}{c}{Position Track} 
    & \multicolumn{4}{c}{Rotation Track} 
    & \multicolumn{3}{c}{6-DoF Track} \\
    \cmidrule(lr){2-4} \cmidrule(lr){5-8} \cmidrule(lr){9-11}
    & Level 0 & Level 1 & Overall 
    & Level 0 & Level 1 & Level 2 & Overall 
    & Position & Rotation & Overall \\
    \midrule
    OpenVLA~\cite{openvla}         & 4.9 & 6.9 & 5.5 & 0.0 & 0.0 & 0.0 & 0.0 & 4.1 & 4.2 & 4.1  \\
    SpatialVLA~\cite{spatialvla}      & 1.4 & 3.4 & 2.0 & 0.0 & 0.0 & 0.0 & 0.0 & 2.5 & 0.6 & 0.6 \\
    SoFar~\cite{sofar}  &73.9 & 69.0 & 72.5 &  15.6   & 30.6   &  36.4   &  25.3   &  \textbf{62.3}  &  17.1   & 12.9  \\
    \textbf{Ours}       & \textbf{74.6} & \textbf{70.7} & \textbf{73.5}
                         & \textbf{26.6} & \textbf{41.7} & \textbf{40.9} & \textbf{35.4}
                         & 56.5 & \textbf{22.9} & \textbf{14.1} \\
    \bottomrule
    \end{tabular}%
    }
    \caption{\textbf{Quantitative Comparison in the Robot Manipulation Task of the Open6DOR V2 Benchmark.} Our method demonstrates better performance than the baselines in most tasks.}
    \label{tab:open6dor_execution}
\end{table*}

\noindent \textbf{Results.}
Tab.~\ref{tab:open6dor_execution} reports the quantitative results on robot manipulation in the Open6DOR V2 benchmark. VLA models rarely work without task-specific fine-tuning. In contrast, our method works effectively without any task-specific tuning. The pose prediction quality is reflected in downstream manipulation performance. SoFar often fails when accurate orientation is required, whereas our method more reliably supports successful execution.
Fig.~\ref{fig:libero} shows the effectiveness of our method. The VLA baselines fail even to grasp the target object. SoFar fails to predict the accurate orientation, causing the marker to get stuck before entering the mug or failing to place the hammer handle upright. In contrast, our method accurately predicts both position and orientation, satisfying the instruction and successfully completing the manipulation. For additional quantitative comparisons with more recent VLAs, see Sec.~\ref{sec:additional_results}.

\begin{table*}[t]
    \centering
    \small
    \setlength{\tabcolsep}{6pt}
    \renewcommand{\arraystretch}{1.18}
    \resizebox{\textwidth}{!}{%
    \begin{tabular}{llcccccccc}
    \toprule
    \multirow{2}{*}[-1.5ex]{Policy} &
    \multirow{2}{*}[-1.5ex]{Training Data} &
    \multicolumn{2}{c}{\makecell{Put Spoon\\on Towel}} &
    \multicolumn{2}{c}{\makecell{Put Carrot\\on Plate}} &
    \multicolumn{2}{c}{\makecell{Stack Green Block\\on Yellow Block}} &
    \multicolumn{2}{c}{\makecell{Put Eggplant\\in Yellow Basket}} \\
    \cmidrule(lr){3-4} \cmidrule(lr){5-6} \cmidrule(lr){7-8} \cmidrule(lr){9-10}
    & &
    \makecell{Grasp\\Spoon} & Success &
    \makecell{Grasp\\Carrot} & Success &
    \makecell{Grasp\\Green Block} & Success &
    \makecell{Grasp\\Eggplant} & Success \\
    \midrule
    OpenVLA~\cite{openvla}      & OXE~\cite{oxe}    & ~4.2 & ~0.0 & 16.7 & ~0.0 & 12.5 & ~4.2 & 16.7 & ~0.0  \\
    SpatialVLA~\cite{spatialvla}   & OXE~\cite{oxe}    & 25.0 & 12.5 & 45.8 & 25.0 & 75.0 & 33.3 & \textbf{79.2} & 62.5 \\
    SoFar~\cite{sofar} & OrienText300K~\cite{sofar} & 62.5 & 58.3 & 79.2 & \textbf{70.8} & 50.0 & 33.3 & 62.5 & ~8.3  \\
    \textbf{Ours} & Training-free & \textbf{62.5} & \textbf{58.3} & \textbf{83.3} & 45.8 & \textbf{50.0} & \textbf{37.5} & 70.8 & \textbf{66.7}  \\
    \bottomrule
    \end{tabular}%
    }
    \caption{\textbf{Quantitative Comparison in the Robot Manipulation Task of the SIMPLER Benchmark.} For each task, we report the final task success rate (“Success”) together with a partial success metric indicating whether the target object was successfully grasped (e.g., “Grasp Spoon”).}
    \label{tab:simplerenv}
\end{table*}

\subsection{Robot Manipulation Evaluation on SIMPLER}
\label{sec:simplerenv}

\noindent \textbf{Dataset.}
We evaluate on the SIMPLER~\cite{simpler} benchmark under the WidowX + Bridge setup, a simulated manipulation setting designed to reflect common real-world robot configurations.
In this setup, we consider four language-conditioned manipulation tasks: \textit{Put Spoon on Towel}, \textit{Put Carrot on Plate}, \textit{Stack Green Block on Yellow Block}, and \textit{Put Eggplant in Yellow Basket}.

\noindent \textbf{Metrics.}
We report both intermediate grasp success rate and final task success rate for each task.
The former measures whether the robot successfully grasps the target object during execution, while the latter measures whether the full instructed goal state is achieved by the end of the rollout.
Reporting both metrics allows us to separately evaluate object acquisition and full task completion, which is particularly informative in manipulation settings.

\noindent \textbf{Baselines.}
We use the same baselines as in Sec.~\ref{sec:open6dor_manip}, including SoFar, OpenVLA, and SpatialVLA. We follow the evaluation protocol used in SoFar.

\begin{figure}[t]
    \includegraphics[width=\linewidth, trim={0cm 0cm 0cm 0cm},clip]{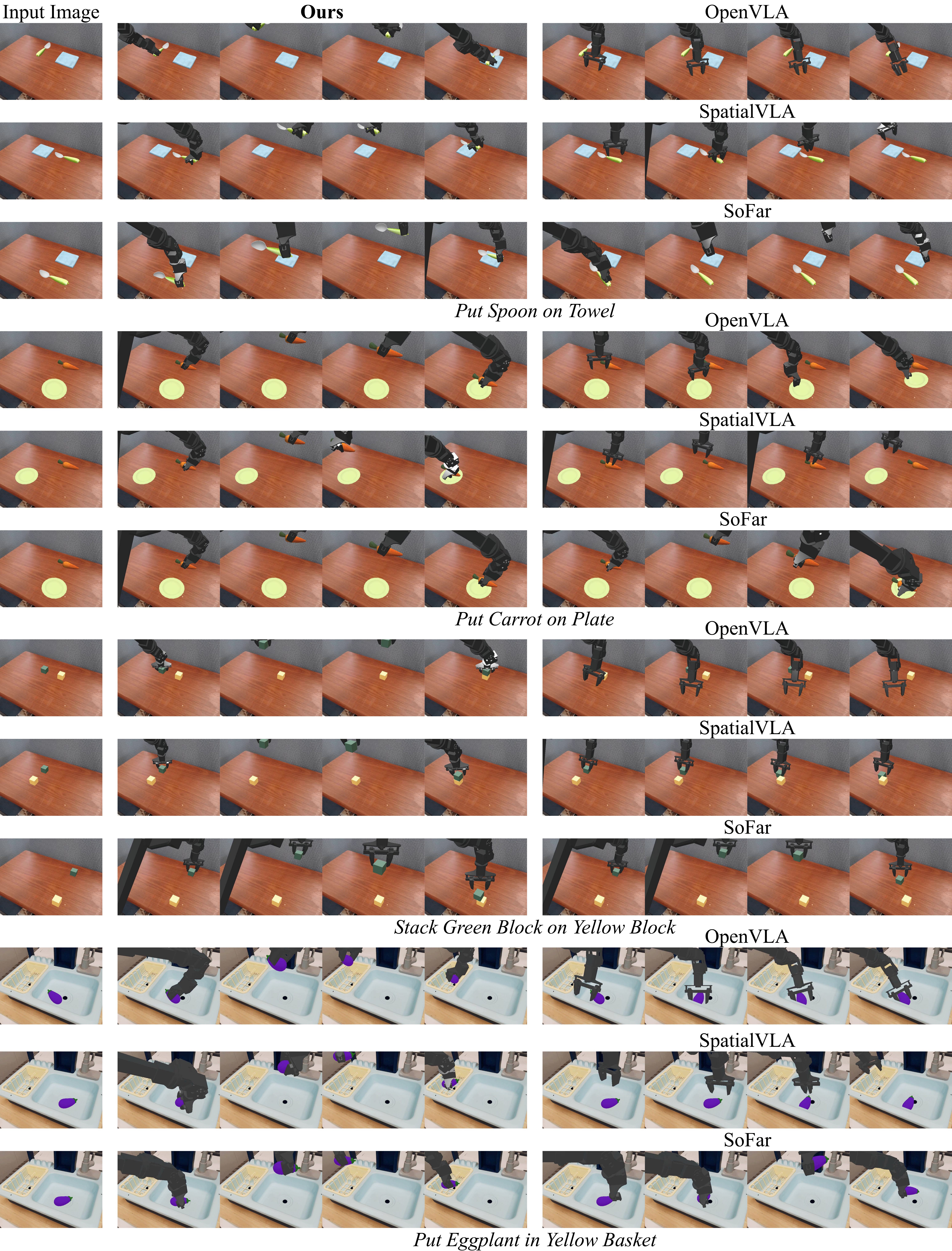}
    \caption{\textbf{Qualitative Comparison in the Robot Manipulation Task of the SIMPLER Benchmark under the WidowX + Bridge Setup.}}
    \label{fig:simpler}
\end{figure}

\noindent \textbf{Results.}
Tab.~\ref{tab:simplerenv} summarizes the quantitative results on SIMPLER. Despite being entirely training-free, our approach achieves competitive performance across the benchmark. In particular, our method achieves the best task success on \textit{Stack Green Block on Yellow Block}, \textit{Put Eggplant in Yellow Basket}, and \textit{Put Spoon on Towel}. Fig.~\ref{fig:simpler} further shows the effectiveness of our method through qualitative examples. These results suggest that iterative visual feedback enables robust downstream robot manipulation even in a fully training-free setting. For additional quantitative comparisons with more recent VLAs, see Sec.~\ref{sec:additional_results}.

\begin{figure}[t]
    \includegraphics[width=\linewidth, trim={0cm 0cm 0cm 0cm},clip]{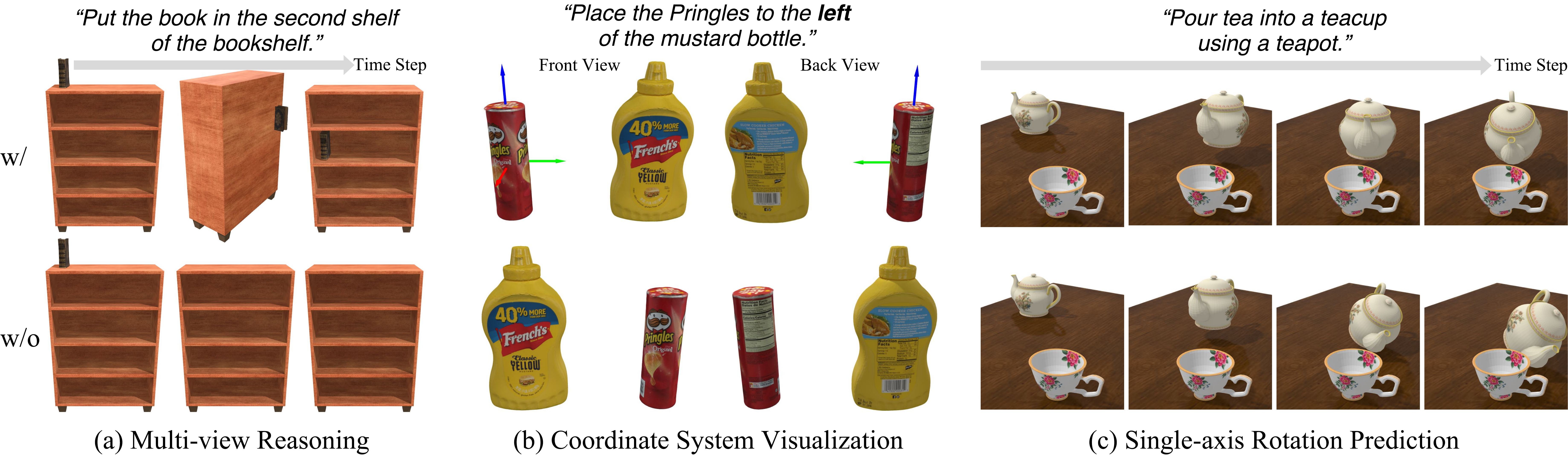}
    \caption{\textbf{Qualitative Ablation Study of Inference-time Techniques.} (a) Without multi-view reasoning, errors caused by occlusion cannot be corrected. (b) Without coordinate system visualization, the VLM struggles to reason consistently about directions across views. (c) Inferring the full SO(3) at once is very difficult for current VLMs.}
    \label{fig:ablation_mesh}
\end{figure}

\begin{table*}[t]
    \centering
    \small
    \setlength{\tabcolsep}{5pt}
    \renewcommand{\arraystretch}{1.15}
    \resizebox{\textwidth}{!}{%
    \begin{tabular}{cccc ccc}
    \toprule
    \multicolumn{4}{c}{Method} 
    & \multicolumn{1}{c}{Position Track}
    & \multicolumn{1}{c}{Rotation Track}
    & \multicolumn{1}{c}{6-DoF Track} \\
    \cmidrule(lr){1-4} \cmidrule(lr){5-5} \cmidrule(lr){6-6} \cmidrule(lr){7-7}
    MV Reasoning & Coord Vis & Single-Axis Rot & Context Memory
    & Overall & Overall & Overall \\
    \midrule
    \cmark  & \cmark & \cmark & 
    & 77.8 & 52.8 & 28.9 \\
    \cmark & \cmark &  & \cmark
    & 80.6 & 57.6 & 36.7 \\
    
    \cmark &  & \cmark & \cmark
    & 71.7 & 56.9 & 31.1 \\
    
     & \cmark & \cmark & \cmark
    & 81.7 & 60.4 & 38.9 \\
    \midrule
    \cmark & \cmark & \cmark & \cmark
    & \textbf{85.6} & \textbf{61.1} & \textbf{40.6} \\
    \bottomrule
    \end{tabular}%
    }
    \caption{Quantitative ablation results on Open6DOR V2 with different combinations of (i) multi-view reasoning, (ii) coordinate system visualization, (iii) single-axis rotation prediction, and (iv) context memory accumulation.}
    \label{tab:ablation}
\end{table*}

\subsection{Ablation Study}
\label{sec:ablation}

\noindent \textbf{Dataset and Metrics.}
We use Open6DOR V2 as in Sec.~\ref{sec:open6dor_6dof}. From each track (position, rotation, and 6-DoF), we randomly sample 180 subtasks and evaluate on a total of 540 subtasks. We report success rate as the evaluation metric, following Sec.~\ref{sec:open6dor_6dof}.

\noindent \textbf{Ablation Settings.}
We ablate the context memory accumulation and the three inference-time techniques described in Sec.~\ref{sec:overview}. Specifically, we evaluate the performance drop caused by removing each of these four components individually. (1) Without multi-view reasoning with supporting view selection, the evaluator is forced to reason from a single input image. (2) Without coordinate system visualization, the coordinate system is described only in text, without any explicit visual overlay. (3) Without single-axis rotation prediction, the VLM is asked to infer 3D Euler angles, which are then applied in x–y–z order. (4) Without context memory accumulation, the VLM does not receive context memory as an additional input.

\noindent \textbf{Results.}
Tab.~\ref{tab:ablation} shows the evaluation results on a subset of Open6DOR V2. Applying all four components yields the best performance across all tasks. We observe that coordinate system visualization and context memory accumulation are particularly important for 6D pose prediction. Coordinate system visualization has a substantial impact on the position track, likely because Open6DOR V2 contains many directional instructions such as front, back, left, and right, and visual cues for the coordinate system play a crucial role in helping the VLM perceive these directions. The effect of removing single-axis rotation prediction is relatively small, likely because most rotation tasks in the Open6DOR V2 benchmark involve only simple target orientations such as upright or lying. In practice, even when instructed to infer 3D Euler angles, the VLM often behaved as though it were performing single-axis rotation prediction, providing a nonzero angle for only one axis while setting the others to zero in most cases. Finally, the effect of multi-view reasoning is relatively small on Open6DOR V2, likely because the benchmark mainly involves simple tabletop placement and the input view already provides sufficient spatial information as shown in Fig.~\ref{fig:open6dor_6dof}.

To further show the importance of each technique, we present qualitative results in Fig.~\ref{fig:ablation_mesh}. In the (bookshelf, book) example, the VLM struggles to recover from severe occlusion without multi-view reasoning. In the (pringles, mustard bottle) example, it becomes confused when directional concepts are specified only in text without coordinate system visualization. In the (teacup, teapot) example, it fails on nontrivial orientations such as \textit{pour} when asked to infer the full SO(3) at once.

\section{Discussion}
\label{sec:discussion}

In this paper, we proposed a framework that uses Vision-Language Models as active closed-loop agents for text-guided 6D object pose rearrangement.
By integrating inference-time techniques such as multi-view reasoning with supporting view selection, object-centered coordinate system visualization, and single-axis rotation prediction, our method effectively mitigates depth ambiguity and the difficulty of full 3D rotation reasoning that plague traditional open-loop, single-view approaches.
Extensive evaluations across multiple benchmarks demonstrate that our framework significantly improves goal 6D pose prediction capabilities and leads to higher success rates in zero-shot robotic manipulation tasks.

Despite these strong results, our approach has a few limitations.
The primary limitation is inference speed.
Operating as a closed-loop system that repeatedly queries a VLM and renders visual feedback at each step incurs a higher computational cost and latency compared to one-shot models.
This makes the current framework less suitable for real-time robotic control scenarios that require low-latency responses.
Additionally, the performance of our system remains inherently limited by the visual understanding ability of the underlying model.
Reducing the computational cost of closed-loop inference and improving the robustness of the underlying VLM are important next steps toward deploying this framework in real-time robotic manipulation.

\section*{Acknowledgements}

This work was supported by RLWRLD, Institute of Engineering Research at Seoul National University, NRF grant funded by the Korean government (MSIT) [No. RS-2022-NR070498, No. RS-2025-25396144, No. RS-2026-25540546, and No. PJT-25-122310], and IITP grant funded by the Korea government (MSIT) [No. RS-2024-00439854, No. RS-2025-25442338, and No. RS-2021-II211343]. H. Joo is the corresponding author.

\clearpage

\bibliographystyle{splncs04}
\bibliography{main}

\clearpage

\appendix
\renewcommand*{\thesection}{\Alph{section}} %

\begin{figure}[t]
    \includegraphics[width=\linewidth, trim={0cm 0cm 0cm 0cm},clip]{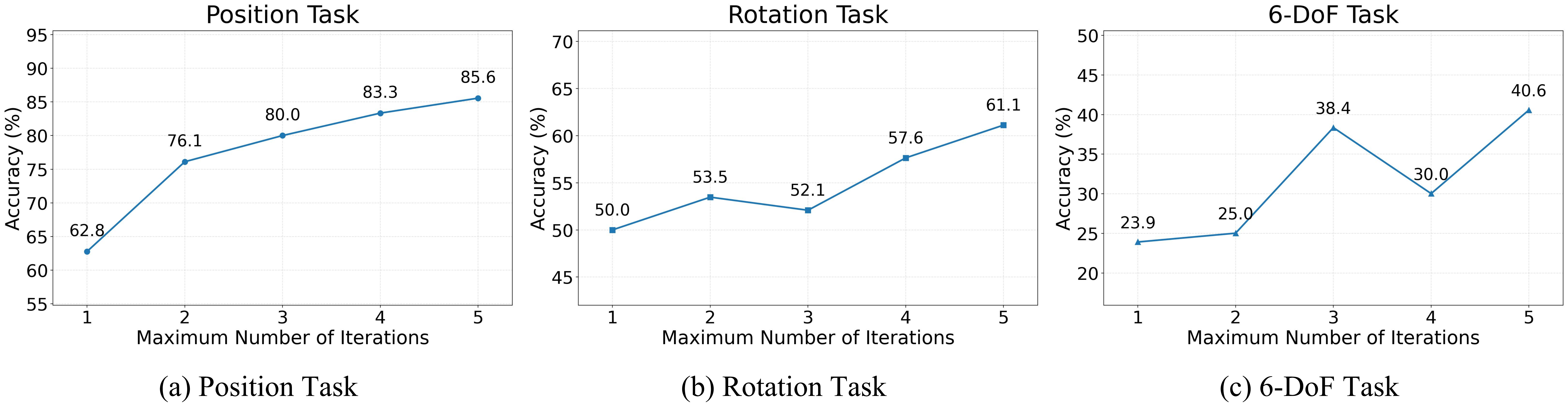}
    \caption{\textbf{Effect of the Maximum Number of Iterations on Open6DOR V2.} Performance generally improves as the maximum number of iterations increases.}
    \label{fig:iter_abl}
\end{figure}

\section{Ablation on the Maximum Number of Iterations}
\label{sec:max_iter_abl}

For relatively simple tasks, such as basic object relocation, the loop often converges within a single iteration, whereas more challenging tasks typically require multiple rounds of reasoning. Moreover, because our method predicts rotations in a single-axis manner, rotational adjustments often require multiple iterations. To address this issue, our method enables iterative reasoning by the VLM while maintaining context memory. This design is based on the assumption that reasoning improves over iterations, which we validate through the following ablation study. The experiments are conducted on the same subset of Open6DOR V2~\cite{sofar, open6dor-gpt} used in Sec.~\ref{sec:ablation} of the main paper, with the maximum number of iterations varying from 1 to 5.

Fig.~\ref{fig:iter_abl} shows the experimental results. The curves exhibit a clear upward trend, indicating that performance generally improves as the maximum number of iterations increases across all tasks. However, a larger maximum number of iterations inevitably incurs greater execution time. Taking the trade-off between performance and inference cost into account, we set the maximum number of iterations to 5 for all experiments in the main paper. In other words, the loop terminates either when the evaluator determines that the current scene is faithful to the text instruction or when the reasoning loop reaches five iterations.

\section{Efficiency Analysis}
\label{sec:efficiency_analysis}

\vspace{-20pt}
\begin{table*}[h]
    \centering
    \scriptsize
    \setlength{\tabcolsep}{3pt}
    \renewcommand{\arraystretch}{1.15}
    \resizebox{\textwidth}{!}{%
    \begin{tabular}{cccc}
    \toprule
    \shortstack[c]{Avg evaluator\\calls/scene} &
    \shortstack[c]{Avg predictor\\calls/scene} &
    \shortstack[c]{Evaluator sec/call\\Mean $|$ p50 $|$ p90} &
    \shortstack[c]{Predictor sec/call\\Mean $|$ p50 $|$ p90} \\
    \midrule
    2.422 & 1.699 & 31.0 $|$ 25.2 $|$ 40.1 & 99.3 $|$ 94.6 $|$ 132.5 \\
    \bottomrule
    \end{tabular}%
    }
    \caption{\textbf{Time/Efficiency Analysis.} We report average evaluator/predictor calls per scene and per-call runtimes in seconds over 1,616 scenes. Runtime statistics are mean, median (p50), and 90th percentile (p90), including rendering time.}
    \label{tab:pos_efficiency_summary}
    \vspace{-20pt}
\end{table*}

We analyze the inference cost of our framework on all 1,616 scenes from the position tasks in Sec.~\ref{sec:open6dor_6dof}. Table~\ref{tab:pos_efficiency_summary} reports the average number of evaluator and predictor calls per scene, together with the per-call runtime of each module measured by the mean, median (p50), and 90th percentile (p90). Our iterative refinement process requires 2.422 evaluator calls and 1.699 predictor calls per scene on average. The predictor accounts for the main runtime bottleneck, taking 99.3 seconds per call on average, while the evaluator takes 31.0 seconds per call. Reducing latency with a lightweight VLM is an important direction for future work.

\begin{figure}[ht]
    \includegraphics[width=\linewidth, trim={0cm 0cm 0cm 0cm},clip]{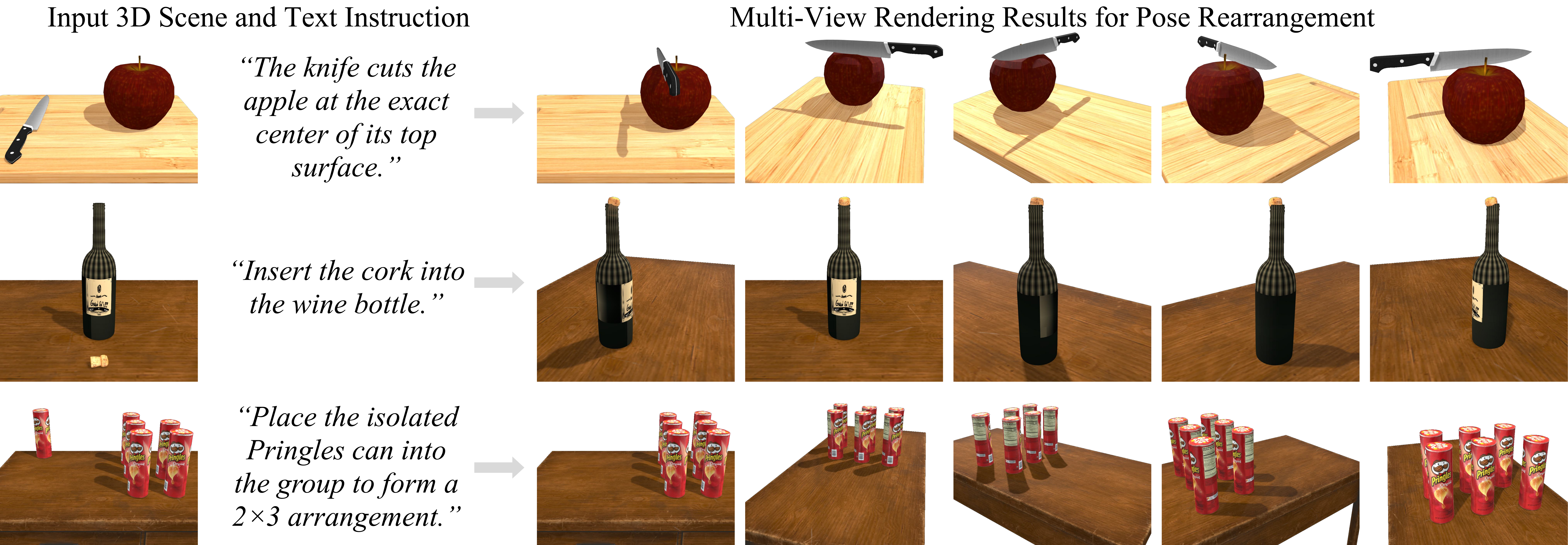}
    \caption{\textbf{Additional Qualitative Results on the 6-DoF Rearrangement Task.} Given an input 3D scene and a text instruction, our method predicts object-level 6D goal poses that satisfy diverse spatial constraints, including precise contact, insertion, and multi-object organization.}
    \label{fig:supp_qual_1}
    \vspace{-30pt}
\end{figure}

\begin{figure}[ht]
    \includegraphics[width=\linewidth, trim={0cm 0cm 0cm 0cm},clip]{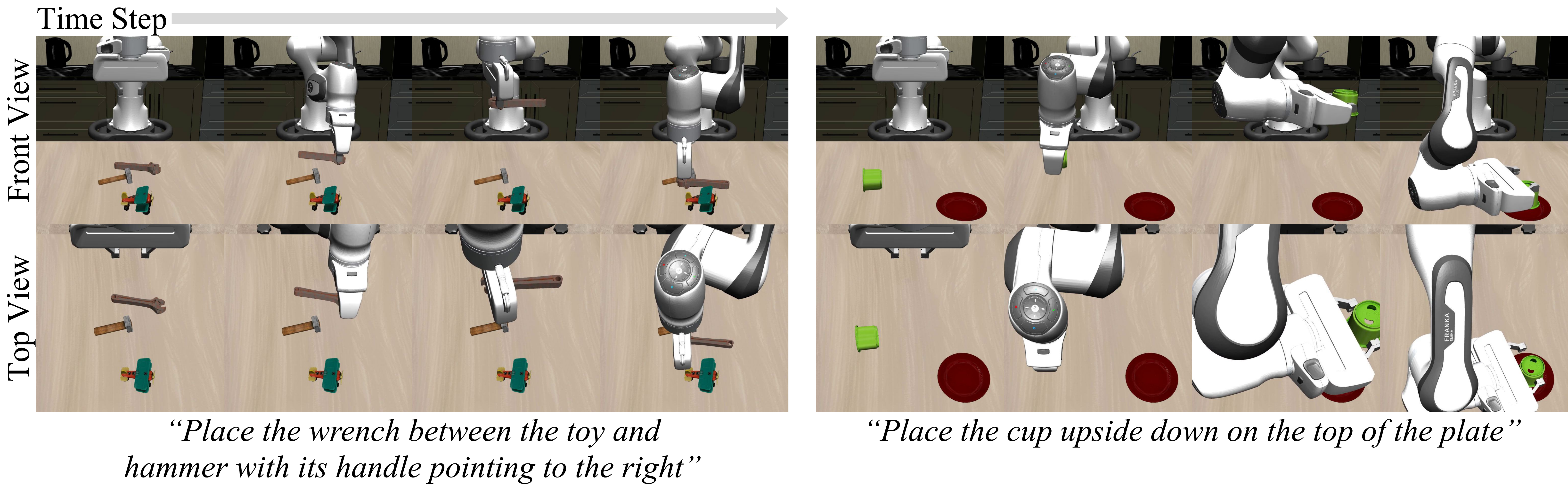}
    \caption{\textbf{Additional Qualitative Results on the Robot Manipulation Task.} We show front-view and top-view rollouts for two orientation-sensitive tasks: placing a wrench between objects with a specified handle direction and placing a cup upside down on a plate. The predicted goal poses translate to successful downstream execution.}
    \label{fig:supp_qual_2}
    \vspace{-30pt}
\end{figure}

\begin{table*}[ht]
    \centering
    \small
    \setlength{\tabcolsep}{3pt}
    \renewcommand{\arraystretch}{1.15}
    \resizebox{\textwidth}{!}{%
    \begin{tabular}{llcccccccccc}
    \toprule
    \multirow{2}{*}{Method} &
    \multirow{2}{*}{\makecell{Fine-tuning\\Data}} &
    \multicolumn{3}{c}{Position Track} &
    \multicolumn{4}{c}{Rotation Track} &
    \multicolumn{3}{c}{6-DoF Track} \\
    \cmidrule(lr){3-5} \cmidrule(lr){6-9} \cmidrule(lr){10-12}
    & &
    Level 0 & Level 1 & Overall &
    Level 0 & Level 1 & Level 2 & Overall &
    Position & Rotation & Overall \\
    \midrule
    $\pi_{0.5}$~\cite{pi05_libero}
    & LIBERO~\cite{libero}
    & 18.4 & 19.3 & 18.7
    & ~6.2 & 16.7 & 22.7 & 13.3
    & 18.2 & ~4.7 & ~2.4  \\

    GR00T~\cite{groot_libero}
    & LIBERO~\cite{libero}
    & ~3.5 & ~5.3 & ~4.0
    & ~0.0 & ~1.4 & ~0.0 & ~0.6
    & ~3.5 & ~1.2 & ~0.6 \\

    \textbf{Ours}
    & --
    & \textbf{74.6} & \textbf{70.7} & \textbf{73.5}
    & \textbf{26.6} & \textbf{41.7} & \textbf{40.9} & \textbf{35.4}
    & \textbf{56.5} & \textbf{22.9} & ~\textbf{14.1} \\
    \bottomrule
    \end{tabular}%
    }
    \caption{\textbf{Additional quantitative comparison on Open6DOR V2.} We compare our method with recent VLA baselines on the position, rotation, and 6-DoF tracks. We use the $\pi_{0.5}$-LIBERO checkpoint for $\pi_{0.5}$ and the GR00T-N1.7-LIBERO checkpoint for GR00T.
    }
    \label{tab:supp_open6dor_execution}
\end{table*}

\begin{table*}[ht]
    \centering
    \small
    \setlength{\tabcolsep}{6pt}
    \renewcommand{\arraystretch}{1.18}
    \resizebox{\textwidth}{!}{%
    \begin{tabular}{llcccccccc}
    \toprule
    \multirow{2}{*}[-1.5ex]{Policy} &
    \multirow{2}{*}[-1.5ex]{Fine-tuning Data} &
    \multicolumn{2}{c}{\makecell{Put Spoon\\on Towel}} &
    \multicolumn{2}{c}{\makecell{Put Carrot\\on Plate}} &
    \multicolumn{2}{c}{\makecell{Stack Green Block\\on Yellow Block}} &
    \multicolumn{2}{c}{\makecell{Put Eggplant\\in Yellow Basket}} \\
    \cmidrule(lr){3-4} \cmidrule(lr){5-6} \cmidrule(lr){7-8} \cmidrule(lr){9-10}
    & &
    \makecell{Grasp\\Spoon} & Success &
    \makecell{Grasp\\Carrot} & Success &
    \makecell{Grasp\\Green Block} & Success &
    \makecell{Grasp\\Eggplant} & Success \\
    \midrule
    $\pi_{0}$~\cite{pi0_simpler}   & BridgeData V2~\cite{bridge}    & \textbf{70.8} & 37.5 & 58.3 & 41.0 & \textbf{66.0} & 20.8 & \textbf{87.5} & \textbf{70.8} \\
    GR00T~\cite{groot_simpler} & BridgeData V2~\cite{bridge} & 45.8 & 45.8 & 41.7 & 25.0 & 37.5 & ~8.3 & 75.0 & 41.7  \\
    \textbf{Ours} & -- & 62.5 & \textbf{58.3} & \textbf{83.3} & \textbf{45.8} & 50.0 & \textbf{37.5} & 70.8 & 66.7  \\
    \bottomrule
    \end{tabular}%
    }
    \caption{\textbf{Additional quantitative comparison on SIMPLER.} 
    We report task success and partial grasp success for each manipulation task. We use the lerobot-pi0-bridge checkpoint for $\pi_0$ and the GR00T-N1.7-SimplerEnv-Bridge checkpoint for GR00T.
    }
    \label{tab:supp_simplerenv}
\end{table*}

\begin{figure}[ht]
    \includegraphics[width=\linewidth, trim={0cm 0cm 0cm 0cm},clip]{Figures/noisy_depth.pdf}
    \caption{\textbf{Behavior with ground-truth and estimated depth.} Our method still predicts a reasonable goal 6D pose using estimated depth from Depth Anything 3 (DA3), but depth errors can shift the perceived object height and lead to an incorrect grasp point.}
    \label{fig:noisy_depth}
\end{figure}

\section{Additional Results}
\label{sec:additional_results}

This section presents additional results. We first provide quantitative comparisons with more recent Vision-Language-Action (VLA) baselines on robot manipulation tasks from Open6DOR V2 and SIMPLER. We then show additional qualitative examples of our method on diverse 6-DoF rearrangement tasks and robot manipulation tasks. Finally, we analyze the behavior of our method when using estimated depth instead of ground-truth depth.

\noindent \textbf{Quantitative Results.}
We conduct additional robot manipulation experiments comparing our method against recent VLA baselines, including $\pi_0$~\cite{pi0_simpler}, $\pi_{0.5}$~\cite{pi05_libero}, and GR00T~\cite{groot_libero, groot_simpler}. Tables~\ref{tab:supp_open6dor_execution} and~\ref{tab:supp_simplerenv} report the results on the Open6DOR V2 and SIMPLER benchmarks, respectively. For each VLA baseline, we use the checkpoint fine-tuned on data that most closely matches the corresponding benchmark environment. As shown in the tables, our method outperforms these baselines in most cases, despite requiring no task-specific fine-tuning.

\noindent \textbf{Qualitative Results.}
We present additional qualitative results obtained with our method. As shown in Fig.~\ref{fig:supp_qual_1}, our framework handles not only simple object relocation but also more structured 6-DoF rearrangement problems that require precise contact, insertion, and multi-object spatial organization. Fig.~\ref{fig:supp_qual_2} further shows that these refined goal poses translate to successful downstream execution even in orientation-sensitive manipulation tasks, where both placement accuracy and object orientation must be satisfied throughout the rollout.

\noindent \textbf{Behavior with Estimated Depth.}
In RGB-D settings, our method uses depth information to lift the scene into 3D, so inaccurate depth estimates can affect downstream execution. Fig.~\ref{fig:noisy_depth} compares results using ground-truth depth and estimated depth from Depth Anything 3~\cite{depthanything3} (DA3). Even with DA3 depth, our method obtains a reasonable goal 6D pose estimate. However, as highlighted in the figure, depth errors can shift the perceived object height, causing the robot to infer an incorrect grasp point. This shows a failure mode of our pipeline when the estimated depth is noisy.

\section{System Prompts}
\label{sec:system_prompt}

In this section, we describe the system prompts used in our framework. Because the prompts in the RGB-D setting are very similar to those in the mesh setting, we focus on the mesh setting for clarity. We use the VLM in three steps: 1) target and related object selection; 2) faithfulness checking by the evaluator; and 3) 6D pose rearrangement prediction by the proposer.

\begin{figure}[t]
    \includegraphics[width=\linewidth, trim={0cm 0cm 0cm 0cm},clip]{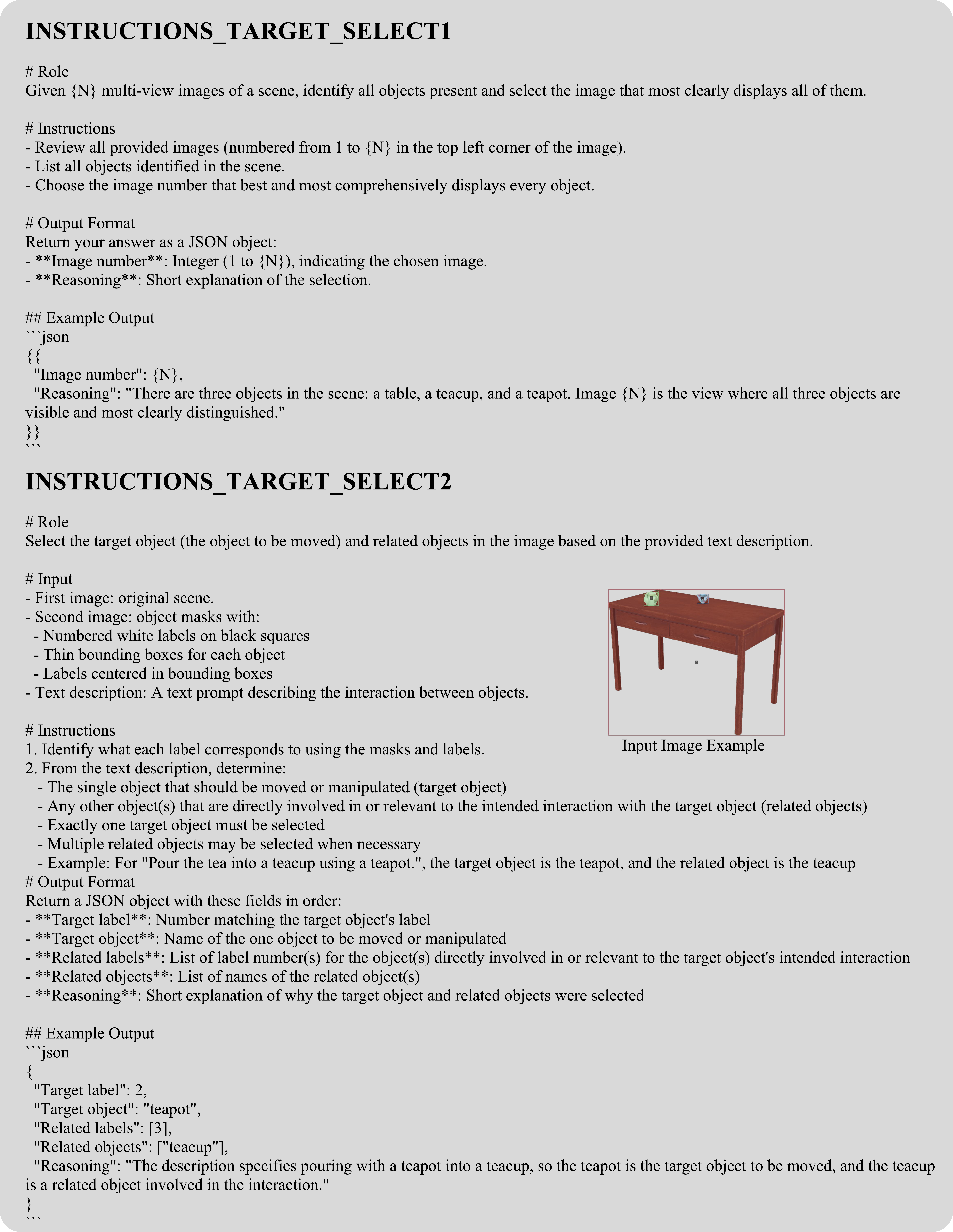}
    \caption{\textbf{System Prompt for Target and Related Object Selection.} This system prompt instructs the VLM to select the view in which the objects are most clearly visible and to identify the target object together with the related objects in that view.}
    \label{fig:system_1}
\end{figure}

\begin{figure}[t]
    \includegraphics[width=\linewidth, trim={0cm 0cm 0cm 0cm},clip]{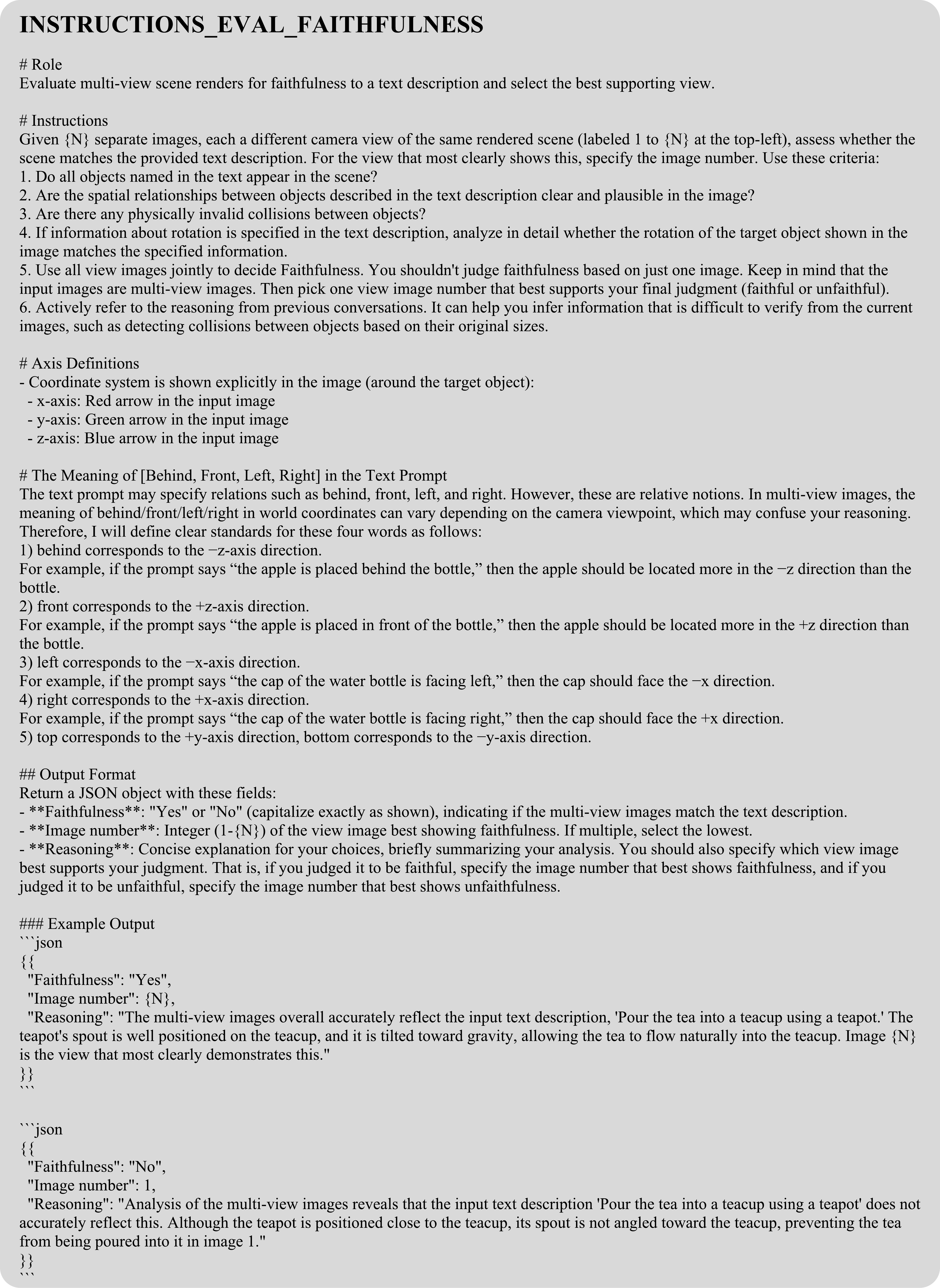}
    \caption{\textbf{System Prompt for Faithfulness Check.} This prompt guides the VLM to assess whether the multi-view images are faithful to the text instruction and to select the view that best supports its judgment.}
    \label{fig:system_2}
\end{figure}

\begin{figure}[t]
    \includegraphics[width=\linewidth, trim={0cm 0cm 0cm 0cm},clip]{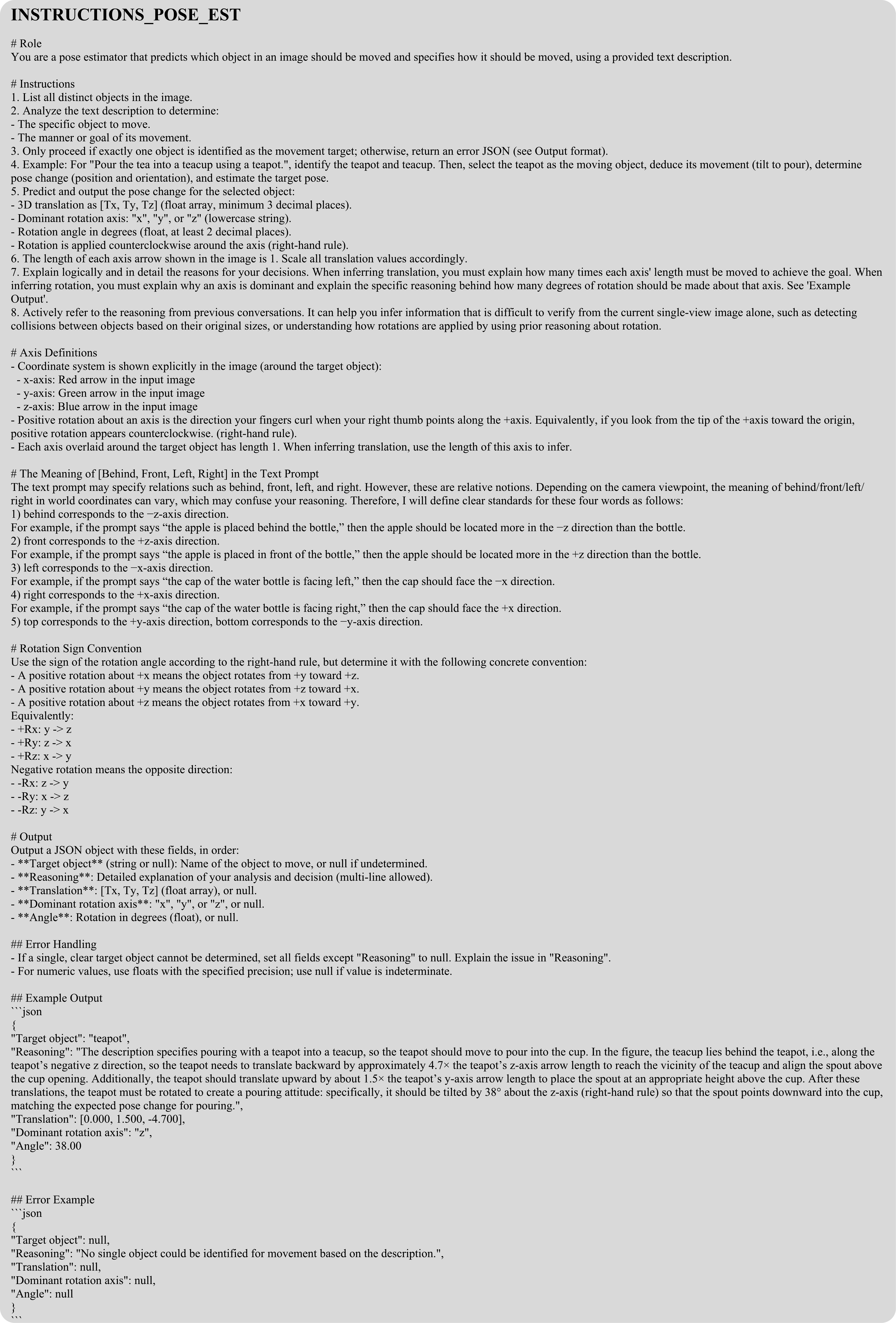}
    \caption{\textbf{System Prompt for 6D Pose Rearrangement Prediction.} This prompt guides the VLM to predict an incremental 6D pose update for the target object.}
    \label{fig:system_3}
\end{figure}

Fig.~\ref{fig:system_1} presents the system prompt for target and related object selection, which instructs the model to select the view in which the objects are most clearly visible and to identify the target object together with the related objects mentioned in the text instruction. Fig.~\ref{fig:system_2} presents the system prompt for faithfulness checking, which instructs the VLM to evaluate whether the input multi-view images are faithful to the text instruction and to provide the view that best supports its judgment. Fig.~\ref{fig:system_3} presents the system prompts for incremental 6D pose update prediction, which instruct the VLM to propose an incremental 6D pose update for the target object so that it aligns with the text instruction.

\end{document}